\documentclass[conference]{IEEEtran}
\IEEEoverridecommandlockouts
\usepackage{cite}
\usepackage{amsmath,amssymb,amsfonts}
\usepackage{algorithmic}
\usepackage{graphicx}
\usepackage{textcomp}
\usepackage{listings}
\usepackage{xcolor}
\usepackage{lipsum}
\usepackage{url}
\usepackage{comment}
\usepackage[ngerman]{babel}

\usepackage{dblfloatfix}  

\definecolor{light-gray}{gray}{0.95}
\lstdefinestyle{mystyle}{
    backgroundcolor=\color{light-gray},  
    basicstyle=\ttfamily\small
}

\usepackage{bookmark}
\usepackage{hyperref}
\hypersetup{
    colorlinks=true,
    linkcolor=black,
    anchorcolor=black,
    citecolor=black,
    filecolor=black,
    menucolor=black,
    runcolor=black,
    urlcolor=black
        }
\renewcommand{\figurename}{\textsc{Figure}}          %
\newcommand{\negspace}{\vspace{-.45cm}}

\usepackage[absolute]{textpos}

\def\BibTeX{{\rm B\kern-.05em{\sc i\kern-.025em b}\kern-.08em
    T\kern-.1667em\lower.7ex\hbox{E}\kern-.125emX}}
    
\bookmark[level=0,page=1]{Conference Paper Title}

  \usepackage{caption}
\DeclareCaptionLabelSeparator{doublepoint}{:\space}
\begin{document}

\title{
Prävention und Beseitigung von Fehlerursachen im Kontext von unbemannten Fahrzeugen
}

\author{
\IEEEauthorblockN{Aron Schnakenbeck$^\ast$, Christoph Sieber, Luis Miguel Vieira da Silva,\\ Felix Gehlhoff}
\IEEEauthorblockA{
    \textit{Institut für Automatisierungstechnik}\\
    \textit{Helmut-Schmidt-Universit\"{a}t/Universit\"{a}t der Bundeswehr Hamburg}\\
    Hamburg, Deutschland \\
    aron.schnakenbeck@hsu-hh.de
    }
\and
\IEEEauthorblockN{Alexander Fay}
\IEEEauthorblockA{\textit{Lehrstuhl für Automatisierungstechnik} \\
\textit{Ruhr-Universität Bochum}\\
Bochum, Deutschland}
}

\maketitle

\begin{abstract}
Unbemannte Fahrzeuge sind durch zunehmende Autonomie in der Lage in unterschiedlichen unbekannten Umgebungen zu operieren. Diese Flexibilität ermöglicht es ihnen Ziele eigenständig zu erfüllen und ihre Handlungen dynamisch anzupassen ohne starr vorgegebenen Steuerungscode. Allerdings erschwert ihr autonomes Verhalten die Gewährleistung von Sicherheit und Zuverlässigkeit bzw. der Verlässlichkeit, da der Einfluss eines menschlichen Bedieners zur genauen Überwachung und Verifizierung der Aktionen jedes Roboters begrenzt ist.
Daher werden Methoden sowohl in der Planung als auch in der Ausführung von Missionen für unbemannte Fahrzeuge benötigt, um die Sicherheit und Zuverlässigkeit dieser Fahrzeuge zu gewährleisten.
In diesem Artikel wird ein zweistufiger Ansatz vorgestellt, der eine Fehlerbeseitigung während der Missionsplanung und eine Fehlerprävention während der Missionsausführung für unbemannte Fahrzeuge sicherstellt. Die Fehlerbeseitigung basiert auf formaler Verifikation, die während der Planungsphase der Missionen angewendet wird. Die Fehlerprävention basiert auf einem regelbasierten Konzept, das während der Missionsausführung angewendet wird.
Der Ansatz wird an einem Beispiel angewendet und es wird diskutiert, wie die beiden Konzepte sich ergänzen und welchen Beitrag sie zu verschiedenen Aspekten der Verlässlichkeit leisten.

\end{abstract}



\section{Einleitung}
Unbemannte Fahrzeuge (Ux\kern-1pt V) werden in den unterschiedlichsten Umgebungen eingesetzt, etwa im Wasser, an Land, in der Luft oder auch im Weltraum. Ihre Vielseitigkeit in Bezug auf Fähigkeiten und Erscheinungsformen eröffnet immer weitere Anwendungsszenarien \cite{Kunze18}. Die kontinuierliche technologische Entwicklung fördert den Einsatz von autonomen Ux\kern-1pt V, die ein zunehmendes Maß an autonomem Verhalten aufweisen. Autonome Ux\kern-1pt Vs zeichnen sich dadurch aus, dass sie in teilweise oder vollständig unbekannten Umgebungen operieren können. Dies ermöglicht den Einsatz von unbemannten Bodenfahrzeugen (UGV) oder unbemannten Luftfahrzeugen (UAV). Diese führen keinen streng vordefinierten Steuerungscode aus, sondern haben die Fähigkeit selbstbestimmt zu erkennen, zu handeln und zu reagieren, um vorgegebene Ziele zu erreichen. Besonders vielversprechend ist die Kombination mehrerer heterogener autonomer Ux\kern-1pt Vs zu einem Verbund. Einzelne Fahrzeuge können sich gegenseitig ergänzen und Schwächen kompensieren, sodass komplexere Missionen ausgeführt werden können \cite{Rizk19}. Selbst für größere Verbünde ermöglicht diese Autonomie die Unabhängigkeit von einem menschlichen Bediener und reduziert den Bedarf an Kontrolle und Überwachung auf ein Minimum.

Ein wichtiger Aspekt zur Sicherstellung der Autonomie eines Systems (hier autonome Ux\kern-1pt Vs) ist dessen Verlässlichkeit (engl. \textit{dependability}), sodass Autonomie nur mit verlässlichen Robotern erreicht werden kann.
Die Verlässlichkeit eines Systems wird von Avizienis et al. \cite{Avizienis04} als Fähigkeit definiert, einen Serviceausfall zu vermeiden, der häufiger und schwerwiegender auftritt, als hinzunehmen ist.
Da die Autonomie autononomer Ux\kern-1pt Vs möglichst wenig menschliche Eingriffe und Überwachung vorsieht, können nicht alle auftretenden Fehler durch menschliches Eingreifen verhindert werden, und es kann nicht auf alle auftretenden Fehler durch einen menschlichen Bediener reagiert werden. 
Daher muss die akzeptable Fehlerhäufigkeit für autonome Ux\kern-1pt Vs als extrem niedrig eingestuft werden, was die Bedeutung der Verlässlichkeit erhöht.

Avizienis et al. \cite{Avizienis04} definieren die Attribute Verfügbarkeit, Zuverlässigkeit (im Sinne von engl. \textit{reliability}), Sicherheit (im Sinne von engl. \textit{safety}), Wartbarkeit und Integrität zur Erreichung der Verlässlichkeit.
Die Attribute \textit{Verfügbarkeit} und \textit{Zuverlässigkeit} sind notwendig, damit ein Ux\kern-1pt V wirklich autonom sein kann. Wenn das Ux\kern-1pt V nicht betriebsbereit oder in der Lage ist, einen korrekten Dienst fortzusetzen, kann keine Autonomie erreicht werden.
\textit{Sicherheit} muss von autonomen Ux\kern-1pt V selbst gewährleistet werden. Das Fehlen von menschlichen Eingriffen, die im Zweifel schwerwiegende Fehler verhindern, muss von den autonomen Ux\kern-1pt V selbst kompensiert werden.
Eine weitere Anforderung an autonome Ux\kern-1pt Vs ist deren \textit{Wartbarkeit}. Reagiert ein autonomes Ux\kern-1pt V auf Umweltveränderungen, bspw. mit Systemanpassungen, so muss sichergestellt werden, dass das Ux\kern-1pt V Wartungen auf autonome Weise durchführen kann. Allerdings muss auch sichergestellt werden, dass die Änderungen durch solche Wartungen nicht zu neuen Fehlerursachen führen und die \textit{Integrität} des Systems verletzen.

Zur Sicherstellung der Vielzahl dieser Attribute, und damit der Sicherstellung der Verlässlichkeit, bedarf es mehrerer, teils sehr verschiedener Methoden. Fokus dieser Arbeit sind die Aspekte der Zuverlässigkeit und Sicherheit, für deren Sicherstellung zwei unabhängige Konzepte vorgeschlagen werden: eine Methode zur Fehlerbeseitigung während der Planungsphase und eine Methode zur Fehlerprävention während der Ausführungsphase einer Mission.

Nach der Beschreibung von Grundlagen zur Missionsplanung für autonome Ux\kern-1pt Vs in Abschnitt~\ref{sec:grundlagen} wird in Abschnitt~\ref{sec:verification} ein formaler Verifikationsansatz vorgestellt. Dieser ermöglicht eine Verifikation von Missionsplänen während der Planungsphase und gegebenenfalls die Beseitigung von gefundenen Fehlerursachen aus den Plänen. Die Missionspläne werden in die Spezifikationssprache GRAFCET transformiert, um sie mittels statischer Analysen und Model Checking untersuchen zu können.
Abschnitt~\ref{sec:ruleBased} präsentiert ein Konzept, um sicherzustellen, dass autonome Ux\kern-1pt Vs während der anschließenden Missionsausführung trotz minimaler menschlicher Überwachung gegebene Sicherheitsvorgaben einhalten, um so die Fehlerprävention sicherzustellen.
Dieser zweistufige Ansatz ist notwendig, da autonome Ux\kern-1pt Vs in einer dynamischen Umgebung operieren und nicht alle möglichen Ereignisse in der Planungsphase abgedeckt werden können.
Der Ansatz wird in Abschnitt~\ref{sec:useCase} auf eine exemplarische Mission angewendet, die von zwei autonomen Ux\kern-1pt Vs ausgeführt wird, bevor in Abschnitt~\ref{sec:discuss} diskutiert wird, wie die beiden vorgestellten Verifikationskonzepte zusammenhängen und welchen Beitrag die Ansätze zur Erreichung der Verlässlichkeit leisten.

\section{Grundlagen}
\label{sec:grundlagen}
Dieser Beitrag betrachtet autonome Ux\kern-1pt V, die sich selbstständig bewegen und in einer unbekannten und unkontrollierbaren Umgebung geeignete Aktionen wählen müssen \cite{ISO8373}.
Der gemeinsame Einsatz von autonomen Ux\kern-1pt V in einem Verbund ermöglicht die Durchführung einer Vielzahl komplexer \emph{Szenarien}.
Im Kontext dieses Artikels beschreibt ein Szenario eine Konstellation von Bedingungen und Umständen, wie die aktuell verfügbaren Ux\kern-1pt V, sowie ein übergeordnetes Ziel, das erreicht werden soll \cite{Sieber.2022}.
Um das übergeordnete Ziel eines Szenarios zu erreichen, werden \emph{Missionen} verwendet. Eine Mission besteht aus einer Abfolge von einem oder mehreren \emph{Missionskommandos} und den erforderlichen Parametern eines Missionskommandos, die einem einzelnen Ux\kern-1pt V zugewiesen werden.
Zum Beispiel kann eine Mission eines autonomen Ux\kern-1pt V nur aus einem Missionskommando \verb|Bewegung| mit den Parametern \verb|pos_x| und \verb|pos_y| der Zielposition in $x$- und $y$-Koordinaten sowie einem Parameter \verb|vel| der Geschwindigkeit, mit der das Ux\kern-1pt V das Ziel ansteuern soll, bestehen.
Diese Missionen werden aus einem Szenario abgeleitet \cite{Sieber.2022}.

Das Ableiten oder Planen von Missionen für einen Verbund autonomer Ux\kern-1pt V für ein bestimmtes Szenario kann sehr komplex werden und zu einer Vielzahl möglicher Lösungen führen, sodass eine automatisierte Planung erstrebenswert ist.
Automatisierte Planung wurde im Bereich der KI-Planung viele Jahre lang untersucht, mit dem Ziel, eine Abfolge von Aktionen (im Kontext dieser Arbeit Missionskommandos) zu finden, die von einem Anfangszustand zu einem gewünschten Zielzustand führen \cite{Russell.2021}.
Die am weitesten verbreitete Sprache im Bereich der KI-Planung ist die \emph{Planning Domain Definition Language} (PDDL) \cite{aeronautiques1998pddl}.
PDDL ist eine domänenunabhängige Sprache, die zur Beschreibung von Planungsproblemen verwendet wird, indem sowohl die Domäne als auch das Problem separat beschrieben werden.
Die Domäne beschreibt hauptsächlich die von jeder Ressource bereitgestellten Aktionen mit ihren Vorbedingungen und Wirkungen.  
Das Problem definiert den Anfangszustand sowie den Zielzustand \cite{aeronautiques1998pddl}.
Zur Lösung eines solchen Planungsproblems werden Planer eingesetzt, die unter anderem Satisfiability Modulo Theories (SMT) verwenden, indem sie die Planungsprobleme als Erfüllbarkeitsprobleme in SMT formulieren und dann Solver verwenden, um sie zu lösen \cite{CMZ_PlanningforHybridSystems_2020}.  
Wenn alle Gleichungen durch Zuweisung von Werten zu den Variablen erfüllt werden können, existiert ein Plan, der vom Startzustand zum Endzustand führt \cite{CMZ_PlanningforHybridSystems_2020}.
Eine Herausforderung bei der Umsetzung solcher Ansätze mit PDDL in der KI-Planung ist der Aufwand, der für die Erstellung eines solchen Planungsproblems erforderlich ist.
Ein weiteres Problem ist, dass der Einsatz von KI-Planung in realen Anwendungen selten ist, da die Ausdruckskraft von PDDL nicht ausreicht \cite{rogalla_improved_2018}.

Darüber hinaus gibt es Ansätze, die eine automatisierte Planung auf Basis von Informationsmodellen anstreben. Ansätze wie in \cite{VieiradaSilvaLuisMiguel.2023} konzentrieren sich darauf, Funktionen von autonomen Ux\kern-1pt V formal zu beschreiben, um die folgenden zwei Aspekte anzugehen: Einerseits kann die Heterogenität verschiedener Ux\kern-1pt V in einem Verbund überwunden werden und einzelne Ux\kern-1pt V in einem Verbund können aufgrund des Informationsmodells bei Bedarf ausgetauscht werden. Andererseits kann die automatisierte Planung erleichtert werden.
Solche Modelle sind komplex und deren Erstellung zeitaufwendig, sodass Ansätze zur automatischen Erstellung eines solchen Modells  wie in \cite{VieiradaSilvaLuisMiguel.91220239152023} vorteilhaft sind.
Es mangelt jedoch noch an Ansätzen, die eine automatisierte Planung basierend auf einem Informationsmodell durchführen.
Bisher lag der Fokus auf dem Informationsmodell und Methoden zur Generierung des Modells, wobei formales Schließen oder KI-Planung für die automatisierte Planung vorgeschlagen wurden.
Erste Ansätze in diese Richtung werden beispielsweise in \cite{Kocher.14.12.2023} vorgestellt, bei denen ein Planungsproblem als Erfüllbarkeitsproblem in SMT aus einem Fähigkeitsmodell automatisch generiert und anschließend gelöst wird.

Aufgrund der Komplexität der Erstellung von Planungsproblemen einerseits und andererseits des Mangels an geeigneten Mitteln für die automatisierte Planung basierend auf formalen Modellen werden Missionen für autonome Ux\kern-1pt V oft noch manuell geplant.
Sowohl manuelle als auch automatisierte Planung führt immer zu einem \emph{Plan}, wie in Listing~\ref{listing:PlanStruktur} gezeigt.
Grundsätzlich umfasst ein Plan eine Abfolge von Aktionen mit den entsprechenden Parametern und stellt eine mögliche Lösung für ein definiertes Problem dar, um das zuvor festgelegte Ziel zu erreichen.
Ein Plan besteht aus verschiedenen Zeitpunkten.
Zu jedem Zeitpunkt können eine oder mehrere Aktionen aufgelistet werden, die zu diesem Zeitpunkt ausgeführt werden sollen.
Wenn mehrere Aktionen einem Zeitpunkt zugeordnet sind, wie in Listing~\ref{listing:PlanStruktur} für den Zeitpunkt~\verb|1| gezeigt, werden diese Aktionen parallel ausgeführt.
Für jede Aktion werden die erforderlichen Parameter angegeben.

\begin{lstlisting}[language=Python, label=listing:PlanStruktur, caption=Allgemeine Struktur eines Plans bestehend aus verschiedenen Zeitpunkten und entsprechenden Aktionen mit den erforderlichen Parametern.]
0: Aktion_x Par_a
1: Aktion_b Par_x Par_c Par_f
   Aktion_c Par_d
2: Aktion_a Par_b
\end{lstlisting}

Missionen werden verwendet, um die Informationen aus einem Plan auf die einzelnen autonomen Ux\kern-1pt V zu übertragen.
Missionskommandos entsprechen den Aktionen und sind eindeutig autonomen Ux\kern-1pt V zugeordnet, sodass eine Mission eines einzelnen autonomen Ux\kern-1pt V aus seinen Missionskommandos in der Reihenfolge besteht, in der sie im Plan vorkommen.
\emph{Bedingungen} sind erforderlich, wenn ein autonomes Ux\kern-1pt V ein Missionskommando erst nach Ausführung eines bestimmten Missionskommandos durch ein anderes Ux\kern-1pt V ausführen soll.
Um fehlerhafte Pläne und damit Fehler in der Missionsausführung zu vermeiden, müssen solche Pläne in der Planungsphase verifiziert werden, was insbesondere für manuell erstellte Pläne gilt.
Jedoch müssen auch automatisch erstellte Pläne je nach gewähltem Ansatz verifiziert werden, da beispielsweise komplexe Einschränkungen nicht immer in PDDL ausgedrückt werden können.
Darüber hinaus müssen selbst korrekte Pläne für autonome Ux\kern-1pt V zur Laufzeit weiter verifiziert werden. Pläne beschreiben nur einen groben Ablauf, da autonome Ux\kern-1pt V in einer unbekannten Umgebung arbeiten. Das bedeutet, dass ihr tatsächliches Verhalten leicht vom Plan abweichen oder ihn erweitern kann, ohne den Zweck des Plans zu gefährden. Autonome Ux\kern-1pt V müssen auf ihre Umgebung reagieren und daher beispielsweise ihren Weg zu einem Wegpunkt selbstständig bestimmen, indem sie Hindernisse durch Anpassung der Koordinaten umgehen.
Dementsprechend müssen zwei Dinge sichergestellt werden:

\begin{enumerate}
\item Pläne müssen formal verifiziert werden, bevor sie an die Ux\kern-1pt V übergeben werden, um Fehler während der Ausführung im Voraus zu verhindern.
\item In der Ausführungsphase sind die Ux\kern-1pt V vielen Dynamiken ausgesetzt, daher muss ihr tatsächliches Verhalten auch während der Ausführung verifiziert werden.
\end{enumerate}

\section{Verifizierungsmethoden}
Dieser zweistufige Ansatz wird im Folgenden vorgestellt.

\subsection{Beseitigung von Fehlerursachen vor Missionsausführung}
\label{sec:verification}
Um sicherzustellen, dass die Pläne fehlerfrei sind, wird formale Verifikation als Methode zur Beseitigung von Fehlerursachen eingesetzt. 
Die Beseitigung von Fehlerursachen wird von Avizienis et al. \cite{Avizienis04} als Mittel zur Reduzierung der Anzahl und Schwere von Fehlerursachen beschreiben.
Luckcuck et~al. \cite{Luckcuck19} präsentieren eine Übersicht, wie formale Methoden im Kontext von autonomen Ux\kern-1pt V eingesetzt werden. Unter anderem werden Ansätze untersucht, die formale Methoden während des Entwurfs von Missionsplänen nutzen, welche später von Verbünden ausgeführt werden, wie z.B. in \cite{Kloetzer11, Hilaire07, Talamadupula14}. Diese Ansätze konzentrieren sich jedoch mehr auf die Erstellung der Spezifikation, anstatt formale Verifikation zur Sicherstellung von Anforderungen einzusetzen.

Um eine formale Verifikation zu ermöglichen, müssen die Pläne in einer formalen Sprache modelliert werden. 
In dieser Arbeit wird GRAFCET \cite{60848} als graphische Modellierungssprache genutzt. GRAFCET wurde ursprünglich für die Modellierung von Steuerungsverhalten im Bereich der industriellen Fertigung entwickelt. 
GRAFCET ist für die Modellierung von Plänen geeignet, da gleichzeitig ablaufende Missionen über nebenläufige Sequenzen und Abhängigkeiten zwischen den Missionen über interne Variablen modelliert werden können.
Zudem ist GRAFCET durch die graphische Repräsentation leicht verständlich und in der Automatisierung weit verbreitet. In der Vergangenheit haben einige Autoren dieses Beitrages bereits Methoden zur Verifikation von GRAFCET untersucht \cite{Mross22, Schnakenbeck23a}, die im Folgenden auf Missionsplänen angewendet werden. Ein erster Schritt ist die Transformation der Missionspläne in GRAFCET. Anschließend werden im GRAFCET-Modell gewünschte Verhaltenseigenschaften verifiziert.
Die zu verifizierenden Eigenschaften müssten dazu formalisiert werden, bei der Anwendung von Model Checking beispielsweise in einer temporalen Logik wie Computation Tree Logic (CTL) \cite{Baier08}. 
Nach Verifikation der Missionspläne können diese von den Ux\kern-1pt Vs ausgeführt werden.

\subsubsection{Transformation von Missionsplänen in GRAFCET}
\begin{figure}
    \centering
    \includegraphics[scale=.75]{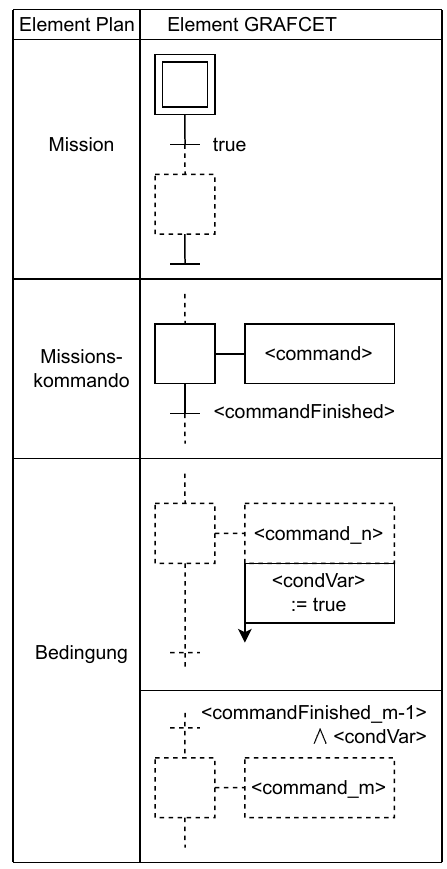}\caption{Regeln zur Transformation eines Plans (linke Seite) in ein GRAFCET-Modell (rechten Seite).}
    \label{fig:transformationRules}
    \negspace
\end{figure}

Die Regeln zur Transformation eines Plans in ein entsprechendes GRAFCET-Modell sind in Abbildung~\ref{fig:transformationRules} dargestellt. Gestrichelte Linien stellen Platzhalterelemente dar, die mittels einer anderen Transformationsregel aus Abbildung~\ref{fig:transformationRules} erzeugt werden. 
Für jede Mission im Plan wird eine neue Sequenz von Schritten (grafisch dargestellt durch ein Quadrat) in dem GRAFCET-Modell generiert, beginnend mit einem Anfangsschritt (doppelt umrandetes Quadrat) und endend mit einer Schlusstransition (grafisch dargestellt durch eine horizontale Linie). Für jeden Missionsbefehl wird ein Schritt in die Sequenz eingefügt, geordnet nach dem Zeitpunkt, an dem er im Plan auftritt. 
Während ein Schritt des GRAFCET-Modells aktiv ist, simuliert eine sogenannte Aktion die Ausführung eines Missionsbefehls \texttt{<command>}. Wenn das Ux\kern-1pt V den Befehl beendet hat, setzt es das Signal \texttt{<commandFinished>} auf \textit{true}, der Schritt wird deaktiviert, und der nächste Schritt wird aktiviert, in dem der nächste Missionsbefehl ausgeführt wird. 

Für jeden Missionsbefehl, der eine Bedingung (\texttt{<command\_m>} in Abb.~\ref{fig:transformationRules}) enthält, wird eine interne boolesche Variable (\texttt{<condVar>}) eingeführt, die anzeigt, ob die Bedingung erfüllt ist. Die vorgelagerte Transitionsbedingung des Schritts, der dem jeweiligen Missionsbefehl entspricht, wird mithilfe eines $\land$-Operators um diese Variable erweitert. 
Die entsprechende Variable wird auf \textit{true} gesetzt, nachdem der Befehl \texttt{<command\_n>} abgeschlossen wurde, der die Bedingung erfüllt.
Dies geschieht mit Hilfe einer sogenannten speichernd wirkenden Aktion, die \texttt{<conVar>} auf \textit{true} setzt, wenn der zugehörige Schritt deaktiviert wird (dargestellt durch den Pfeil).
\begin{figure*}[b]
    \centering
\includegraphics[width=\linewidth]{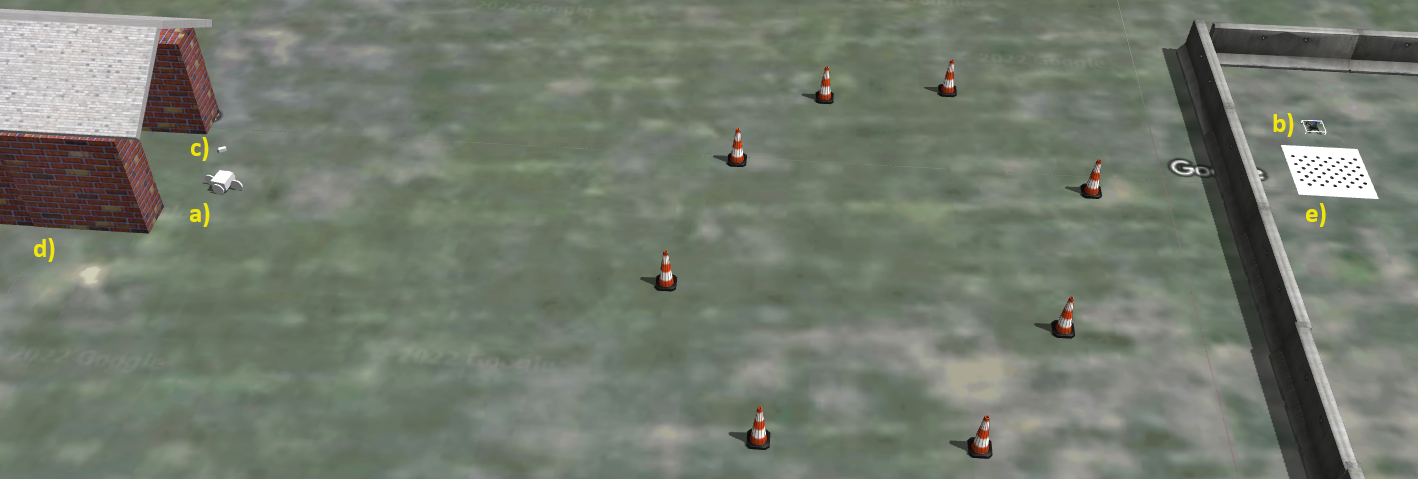}
    \caption{Situation des betrachteten Anwendungsfalls mit a) UGV, b) UAV, c) Paket, d) Fabrikhalle und e) Zielposition.}
    \label{fig:useCaseScenario1}
\end{figure*}
Die Anwendung der Transformationsregeln führt zu einem GRAFCET-Modell, das eine formale Darstellung des Plans ist und verifiziert werden kann.

\subsubsection{Eigenschaften von Missionsplänen}
Um die GRAFCET-Modelle zu verifizieren, müssen Eigenschaften definiert und vom Anwender verifiziert werden:
\begin{itemize}
    \item Strukturelle Anforderungen: Es könnte verlangt werden, dass Missionen verschiedene Arten von Befehlen enthalten, wie das Greifen eines Objekts, das Fahren zu einem Ort oder das Ablegen eines Objekts. Ebenso könnte gefordert werden, dass Missionen mit einem Startbefehl beginnen und mit einem Stoppbefehl enden. Diese Eigenschaften werden mit Hilfe einer graphbasierten Analyse überprüft, die eine Tiefensuche nach Schritten durchführt, die solchen Missionsbefehlen entsprechen.
    \item Chronologische Reihenfolge der Befehle: Einige Arten von Befehlen hängen logisch von anderen ab, z. B. muss auf ein Greifen das Ablegen eines Objektes folgen. Diese Eigenschaften können mit Hilfe von temporalen Logiken formalisiert und mittels Model Checking \cite{Mross22} verifiziert werden. Die Formalisierung solcher Eigenschaften in temporalen Logiken erfordert ein höheres Maß an Fachwissen. Da jedoch jede Mission aus modularen und wiederverwendbaren Missionsbefehlen besteht, ist es möglich, eine bibliotheksartige Liste von Eigenschaften zu definieren, die wiederverwendet werden können.
    Eine solche Befehlsfolge kann in CTL als $AG(\psi_1\rightarrow AF\psi_2)$ formalisiert werden, was bedeutet, dass $\psi_2$ zu einem Zeitpunkt nach $\psi_1$ eintreten muss.
    \item Freiheit von Deadlocks: Wenn zwei Ux\kern-1pt Vs gegenseitig auf die Beendigung einer bestimmten Aufgabe warten, um fortzufahren, könnte dies zu einem Deadlock führen. 
    Mögliche Deadlocks können ebenfalls durch Model Checking erkannt werden.
	\item Abwesenheit von sicherheitskritischen Situationen: Es muss sichergestellt werden, dass bestimmte Befehle nicht gleichzeitig ausgeführt werden können, z. B. dass Ux\kern-1pt Vs nicht zur gleichen Zeit den gleichen Gegenstand greifen oder dass sie nicht gleichzeitig zum selben Ort fahren. Da die Ux\kern-1pt Vs die Befehle selbständig ausführen, hängt ihr genaues Verhalten von den Implementierungsdetails ab. Solche Situationen können zu einem Deadlock führen, z. B. wenn zwei Ux\kern-1pt Vs versuchen, einen Zielort zu erreichen, aber gleichzeitig eine Kollision vermeiden wollen. Um diese Art von Situationen auf Missionsplanebene zu erkennen, können die entsprechenden Eigenschaften entweder mit Hilfe von Model Checking analysiert werden, oder es kann eine statische Analyse verwendet werden, wie in \cite{Schnakenbeck23a} vorgeschlagen.

    Für das Model Checking können diese Situationen mithilfe von Invarianten nachgewiesen werden. Für einen Beispielplan, der zwei Missionsbefehle enthält, die ein Greifen desselben Objekts vorsehen, muss sichergestellt werden, dass die korrespondierenden Schritte im GRAFCET-Modell nicht gleichzeitig aktiv sein können: $AG\neg(step\_1 \land step\_2)$.

\end{itemize}
\begin{figure*}[b]
    \centering
    \includegraphics[width=\textwidth]{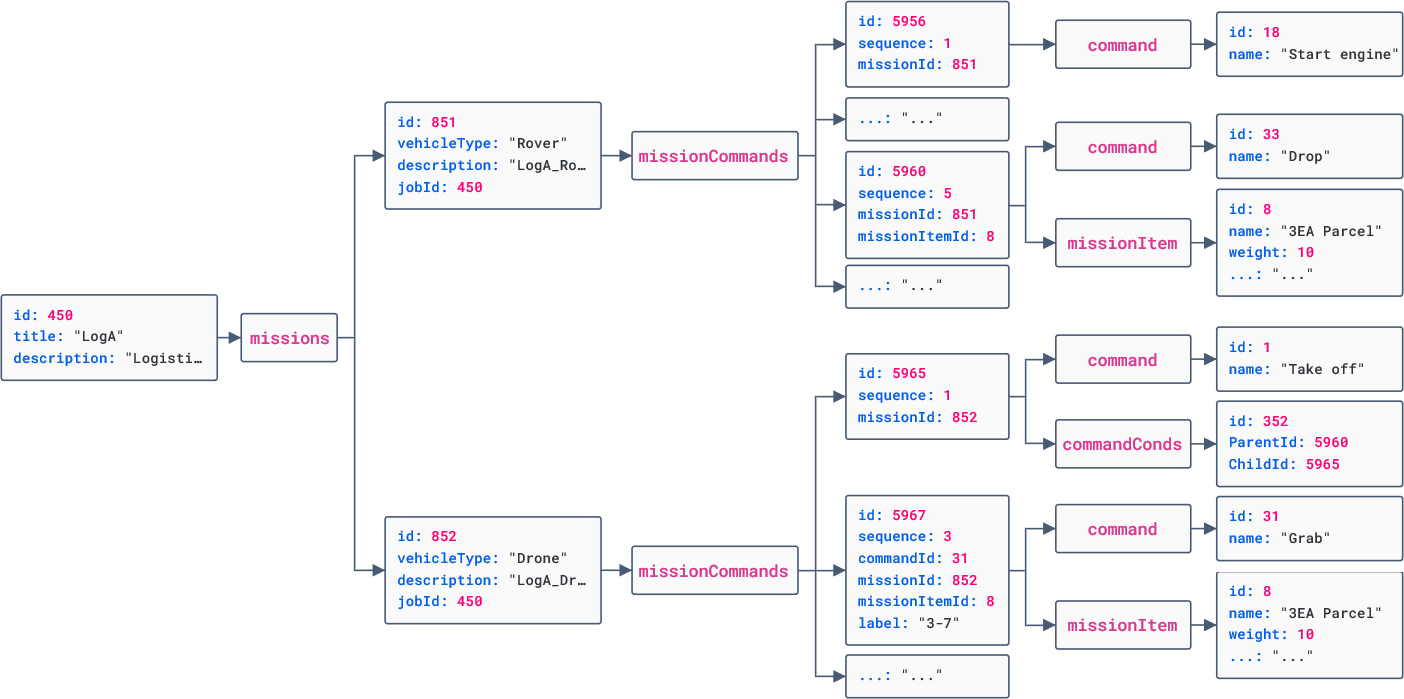}
    \caption{Auszug des Missionsplans im JSON-Format für das Szenario in Abbildung.~\ref{fig:useCaseScenario1}.}
    \label{fig:jsonUseCase}
\end{figure*}

\subsection{Fehlerprävention während der Missionsausführung}
\label{sec:ruleBased}
Nach Avizienis et al. \cite{Avizienis04} umfasst Fehlerprävention zum einen die Vermeidung der Entstehung von Fehlern und zum anderen die Vermeidung der weiteren Auswirkungen bestehender Fehler.
Der im vorangegangenen Kapitel erstellte und verifizierte Missionsplan räumt den einzelnen autonomen Ux\kern-1pt Vs bewusst maximale Handlungsfreiheit ein. Dies hat zur Folge, dass die Missionsausführung nicht vollständig vorhersehbar und daher nur bedingt kontrollierbar ist. Erschwert wird dies zusätzlich durch unbekannte und sich ändernde Umgebungen, z. B. ist die Lage sicherheitsrelevanter Gebiete im Voraus nicht bekannt und kann sich zudem verändern. 
Sind solche Gebiete außerdem nicht direkt relevant für das Missionsziel, ist es möglich, dass ein autonomes Ux\kern-1pt V, trotz korrekter Wahrnehmung des Gebietes, zugehörige Informationen nicht verarbeitet. Wird z. B. ein Sperrgebiet lediglich wahrgenommen, ohne daraufhin den eigenen Pfad entsprechend anzupassen, kann dies zu unsicherem Verhalten führen. Daher ist es ratsam, den Ux\kern-1pt V parallel zum Missionsplan auch Missionsauflagen zu übermitteln. Diese Auflagen legen verbotene, hier sicherheitskritische, Verhaltensweisen fest. Eine missionsunspezifische Formulierung erleichtert die Wiederverwendung von Auflagen \cite{Sieber23Aut}.
In \cite{Sieber24} stellen die Autoren einen regelbasierten Ansatz zur Laufzeitverifikation vor, mit dem sie unsicheres Verhalten von unbemannten Luftfahrzeugen (UAV) erkennen konnten. 
Es wurden jedoch nur Sicherheitsverstöße erkannt und gemeldet, die bereits aufgetreten waren. Eine Meldung veranlasste dann einen menschlichen Nutzer Gegenmaßnahmen einzuleiten. 

Das hier vorgestellte Konzept zur missionsbegleitenden Fehlerprävention greift den Ansatz von \cite{Sieber24} auf und erweitert ihn um eine einerseits präventiv erkennende und andererseits selbstständig reagierende Komponente. Um die Entstehung eines Fehlers mit Hilfe eines regelbasierten Systems wirksam zu verhindern, sind vier Schritte notwendig: (I) Formulierung des Fehlers, (II) Formulierung des drohenden Fehlers, (III) Fähigkeit, den Fehler aus der aktuellen Situation vorherzusagen, (IV) Gegenreaktion auf den drohenden Fehler. Diese vier Schritte werden im Folgenden anhand eines einfachen Beispiels erläutert. Zu diesem Zweck werden ein unbemanntes Bodenfahrzeug (UGV) und ein Sperrgebiet betrachtet. Das UGV darf das Sperrgebiet nicht befahren. Der zugehörige Fehler wird wie folgt formuliert: WENN die Position des UGV im Sperrgebiet liegt, DANN liegt ein Fehler vor (I). Die aktuelle Position des UGV ist somit die Fehlerursache. Ein drohender Fehler kann nun durch die Betrachtung der bevorstehenden Position formuliert werden. WENN die bevorstehende Position des UGV im Sperrgebiet liegt, DANN liegt ein drohender Fehler vor (II). Durch die bewusste Verwendung der Variable \textit{bevorstehende Position} innerhalb der Regel kann der zugehörige Wert auf mehrere Arten ermittelt werden. Im besten Fall sind dem UGV bereits Informationen über zukünftige Ziele und Wegpunkte bekannt. Im ungünstigsten Fall muss die bevorstehende Position anhand der aktuellen Position, des Kurses und der Geschwindigkeit für einen zu wählenden Zeithorizont berechnet werden, z. B. mit einer separaten Regel (III). 
Der Zeithorizont sollte so festgelegt werden, dass eine angemessene Reaktionszeit sowie Zeit für eine Gegenreaktion (z. B. Abbremsen) berücksichtigt wird. Allerdings kann aufgrund der relativ langen Reaktionszeit des Menschen trotz rechtzeitiger Benachrichtigung ein Fehler auftreten, bevor eine Gegenmaßnahme eingeleitet wurde. Daher kann es sinnvoll sein, am Ende der in (II) formulierten Regel bereits eine Gegenreaktion zu definieren: WENN die bevorstehende Position des UGV im Sperrgebiet liegt, DANN liegt ein drohender Fehler vor UND das UGV reduziert seine Geschwindigkeit um 50 Prozent (IV). Die hier exemplarisch vorgeschlagene Geschwindigkeitsreduzierung ermöglicht es dem UGV, einen alternativen Pfad zu bestimmen. Wenn auch dieser Pfad das Sperrgebiet kreuzt, wird die Geschwindigkeit so lange weiter reduziert, bis das UGV beim Einfahren in das Gebiet zum Stillstand kommt. Es ist zweckmäßig, dass in \cite{Sieber24} vorgeschlagene Konzept der Benachrichtigung des menschlichen Nutzers auch bei automatischen Gegenreaktionen beizubehalten, um auch ein menschliches Eingreifen grundsätzlich zu ermöglichen.

\section{Umsetzung und Implementierung}
\label{sec:useCase}
In diesem Abschnitt wird das Potenzial des Ansatzes und das Zusammenspiel der beiden vorgestellten Konzepte anhand eines einfachen Anwendungsfalles demonstriert. Die entsprechende Implementierung wurde mit der in \cite{Sieber.2022} vorgestellten, ROS2-basierten Systemarchitektur durchgeführt. Diese ermöglicht die Trennung von Missionsplanung und \text{-ausführung}, ohne sich im Detail mit dem Aufbau, der Sensorik oder Aktorik der einzelnen Ux\kern-1pt Vs auseinandersetzen zu müssen.

Im gewählten Anwendungsfall wird ein Ux\kern-1pt V-Verbund, bestehend aus einem UGV und einem UAV, verwendet, um den Transport eines Pakets durchzuführen. Dieses Paket befindet sich zunächst an einer Startposition in einer Fabrikhalle. Sein Zielort ist von einem Zaun umgeben. Da weder das UGV noch das UAV allein in der Lage sind, das Paket zu transportieren, muss das Paket außerhalb des eingezäunten Bereichs übergeben werden. Abbildung~\ref{fig:useCaseScenario1} zeigt die Ausgangssituation des Anwendungsfalls. Das UGV befindet sich neben dem Paket. Außerhalb befindet sich das UAV neben der Zielposition. 

Das erwartete Verhalten des Verbunds besteht darin, den Transport des Pakets selbstständig durchzuführen. Der Missionsplan, der dieses Verhalten repräsentiert, wurde manuell über eine Weboberfläche geplant, die den Plan in einem JSON-Format speichert. Die resultierende JSON-Datei für das Beispielszenario ist auszugsweise in Abb.~\ref{fig:jsonUseCase} in einer Graphdarstellung gezeigt. Der Plan umfasst zwei Missionen: Eine für das UGV (oben) und eine für das UAV (unten). Jede Mission besteht aus einer ähnlichen Abfolge von mehreren Missionsbefehlen: \textit{start/takeOff $\rightarrow$ drive/flyTo $\rightarrow$ grab $\rightarrow$ drive/flyTo $\rightarrow$ drop $\rightarrow$ drive/flyTo $\rightarrow$ stop/land}. Da das Paket jedoch zuerst vom UGV transportiert werden muss, wartet das UAV, bis das UGV das Paket an einer Übergabeposition abgesetzt hat. 
Daher wird dem \textit{takeOff}-Befehl der UAV eine Befehlsbedingung hinzugefügt: Das UAV kann erst abheben, wenn das UGV das Paket abgelegt hat.

\subsection{Verifikation der Missionspläne}
\begin{figure}
    \centering    \includegraphics[scale=.75]{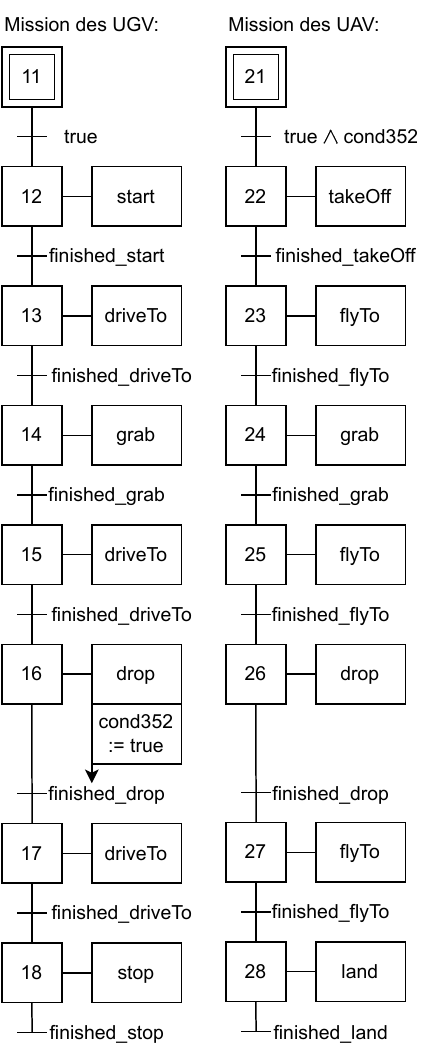}
    \caption{GRAFCET-Modell des Plans für den Anwendungsfall.}
\label{fig:grafcetUseCase}
   \negspace
\end{figure}

Für die Verifikation des Plans während der Entwurfsphase wird der in Abschnitt~\ref{sec:verification} vorgestellte Ansatz auf den erstellten Plan angewendet, der in Abbildung~\ref{fig:jsonUseCase} dargestellt ist. Die Anwendung der GRAFCET-Transformationsregeln resultiert in einem GRAFCET-Modell, das in Abbildung~\ref{fig:grafcetUseCase} dargestellt ist.  

Exemplarisch werden zwei Eigenschaften verifiziert: 
Erstens muss sichergestellt werden, dass der Plan nicht erfordert, dass sich die Ux\kern-1pt Vs näher als in einem bestimmten Sicherheitsabstand zueinander bewegen. Die Formalisierung dieser Eigenschaften kann durch Informationen aus der JSON-Datei aus Abbildung~\ref{fig:jsonUseCase} unterstützt werden, in der die Zielkoordinaten z. B. für die \textit{driveTo} und \textit{flyTo} Befehle gespeichert sind. Wenn die Zielkoordinaten für Befehle verschiedener Ux\kern-1pt Vs übereinstimmen, dürfen die entsprechenden Schritte im GRAFCET-Modell nicht gleichzeitig aktiv sein.
Für das GRAFCET-Modell in Abbildung~\ref{fig:grafcetUseCase} bedeutet dies, dass die Schritte 13, 14 nicht gleichzeitig mit den Schritten 23, 24 aktiv sein dürfen.
In CTL kann das mit den Invarianten $AG \neg (step\_13 \land step\_23)$, $AG \neg (step\_14 \land step\_23)$, usw. formuliert werden.
Zweitens muss sichergestellt werden, dass der Plan durch die Modellierung von Bedingungen keine Verklemmung hervorruft. 
Für jede induzierte Bedingungsvariable (in diesem Fall nur \textit{cond352}) kann mit Hilfe der CTL-Formel $EF( \textit{cond352} == \textit{true})$ überprüft werden, ob sie schließlich erfüllt ist.
Wie in \cite{Mross22} vorgestellt, wurde das GRAFCET-Modell in ein Transitionssystem überführt und mit dem Model Checker ITS-Tools\footnote{\url{https://lip6.github.io/ITSTools-web/}} verifiziert, wodurch die formalisierten Eigenschaften nachgewiesen werden konnten.

\subsection{Laufzeitverifikation der Missionsausführung}
Die mit GRAFCET verifizierten und fehlerfreien Missionspläne  können nun in ihrer ursprünglichen Form (vgl. Abb.~\ref{fig:jsonUseCase}) an UGV und UAV übermittelt werden. Sie gewähren den Ux\kern-1pt Vs maximale Handlungsfreiheit bei der Ausführung ihrer Missionen. 
Wie in der Einleitung erwähnt, muss aufgrund der Autonomie der Ux\kern-1pt Vs sichergestellt werden, dass während der Missionsausführung keine Fehler auftreten. Exemplarisch wird hier die Gefahr durch das Eindringen in Sperrgebiete als Fehlerursache betrachtet. Insbesondere im Zusammenhang mit UAVs werden Sperrgebiete (ugs. Geofence) zur Einhaltung von Abständen, z. B. zu Objekten und Einrichtungen, verwendet \cite{Stevens.2018}. Im Kontext autonomer Ux\kern-1pt V ist es nicht zweckmäßig, feste Pfade festzulegen, um Sperrgebiete zu vermeiden. Stattdessen vermeiden autonome Ux\kern-1pt V diese Sperrgebiete effektiv, indem sie ihre Pfade selbstständig bestimmen. Abbildung~\ref{fig:useCaseScenario1} zeigt ein Sperrgebiet, hervorgehoben durch Pylonen. Es werden zwei einfache regelbasierte Auflagen erstellt, um sicherzustellen, dass weder das UGV noch das UAV in dieses Gebiet eindringen. Analog zu dem Ansatz in \cite{Sieber24} werden diese Regeln innerhalb einer Ontologie mit der SWRL-Regelsprache \cite{SWRL} modelliert und zur Laufzeit ausgewertet. Tabelle~\ref{tab:my_table1} und Tabelle~\ref{tab:my_table2} zeigen zwei SWRL-Regeln. 
Regel~1 in Tabelle~\ref{tab:my_table1} erkennt vorausschauend, ob das UGV in das Sperrgebiet einfährt, und reduziert die Geschwindigkeit, um gegebenenfalls eine Neuplanung des Pfades zu ermöglichen. Wenn das UGV dennoch in das Sperrgebiet einfährt, weil es seine Geschwindigkeit nicht rechtzeitig weit genug reduziert hat oder keine Gegenreaktion erfolgte, wird es durch Regel~2 in Tabelle~\ref{tab:my_table2} gestoppt, indem seine Geschwindigkeit auf null gesetzt wird. 
\begin{table}[t]
    \centering
        \caption{Die Regel R1 verlangsamt das UGV, wenn es in ein Sperrgebiet einfahren wird.}
    \label{tab:my_table1}
    \begin{tabular}{|c|p{6.75cm}|}
        \hline
        \textbf{Line No.} & \textbf{SWRL-Atom} \\
        \hline
        1 & UGV(?myUGV) \\
        2 & \^{} hasImpendingPosition(?myUGV, ?position) \\
        3 & \^{} RestrictedArea(?RA) \\
        4 & \^{} iswithin(?position, ?RA) \\
        5 & \^{} Velocity(?myUGV, ?velocity) \\
        6 & $\rightarrow$ ImpendingFault(?myUGV, “The UGV may enter a restricted area”) \\
        7 & \^{} Velocity(?myUGV, ?newvelocity) \^{} swrlb:multiply(0.5 , ?velocity, newvelocity) \\
        \hline
    \end{tabular}
\end{table}
\begin{table}[t]
    \centering
        \caption{Die Regel R2 bewirkt, dass das UGV anhält, sobald es sich im Sperrgebiet befindet.}
    \label{tab:my_table2}
    \begin{tabular}{|c|p{6.75cm}|}
        \hline
        \textbf{Line No.} & \textbf{SWRL-Atom} \\
        \hline
        1 & UGV(?myUGV) \\
        2 & \^{} hasPosition(?myUGV, ?position) \\
        3 & \^{} RestrictedArea(?RA) \\
        4 & \^{} iswithin(?position, ?RA) \\
        5 & $\rightarrow$ Fault(?myUGV, “The UGV is within a restricted area”) \\
        6 & \^{} Velocity(?myUGV, 0.0) \\
        \hline
    \end{tabular}
\end{table}
    \begin{figure}[t]
    \centering
    \includegraphics[width=\linewidth]{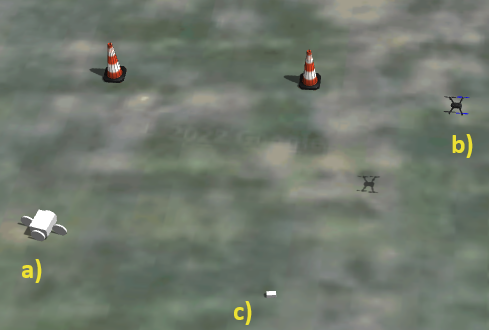}
    \caption{Übergabe des c) Pakets. a) UGV kehrt in die Fabrikhalle zurück und b) UAV nähert sich der Übergabeposition.}
    \label{fig:useCaseScenario2}
       \negspace
\end{figure}

Für Regel~1 verlangt SWRL, dass die aktuelle Geschwindigkeit \textit{?velocity} bereits in Zeile~5 in der Prämisse der Regel abgefragt wird, damit die neue Geschwindigkeit \textit{?newvelocity} in der Konklusion berechnet werden kann. Da Regel~2 die Geschwindigkeit nicht berechnet, sondern fest zuweist, ist hier eine vorherige Abfrage nicht notwendig. Das UAV erhält ähnliche Regeln, die die Geschwindigkeit reduzieren oder einen Schwebeflug (engl. Loiter) bewirken. 
Während der Ausführung der Missionen aktualisieren beide Ux\kern-1pt Vs laufend ihre jeweilige Wissensbasis und analysieren die Einhaltung ihrer jeweiligen Regeln.

Durch die Kombination der beiden Konzepte wird sichergestellt, dass der Packstücktransport kohärent strukturiert ist und keine sicherheitsrelevanten Störungen bei der Durchführung des Einsatzes auftreten. Das UGV holt das Paket zunächst aus der Werkshalle und transportiert es zur Übergabeposition. Die Sicherheitsauflagen verhindern einen geradlinigen Transport durch das Sperrgebiet. Sobald das Paket an der Übergabeposition abgeladen ist, startet das UAV und nähert sich, um das Paket zu übernehmen, während das UGV zurückkehrt. Abbildung 5 zeigt diese Szene im Rahmen der Mission. Nachdem das UAV das Paket am Zielort abgeladen hat, ist die Mission erfolgreich beendet.

\section{Fazit}
\label{sec:discuss}
In diesem Artikel wurden Aspekte der Verlässlichkeit für autonome Ux\kern-1pt V untersucht. Aus der angestrebten Freiheit in der Missionsausführung und der Vielfalt von Ux\kern-1pt V ergeben sich besondere Anforderungen an den Umgang mit Fehlern. Diese Dualität von Selbstständigkeit einerseits und der Notwendigkeit der wirksamen Kontrolle andererseits erschwert die Verifikation sowohl im Rahmen der Missionsplanung als auch der Missionsausführung. Die Teilaspekte der Verlässlichkeit, insbesondere Zuverlässigkeit und Sicherheit, wurden mit Konzepten zur Fehlerbeseitigung und Fehlerprävention behandelt. 
Dennoch sind weder die Fehlerbeseitigung vor der Mission noch die Fehlerprävention während der Mission allein in der Lage, fehlerfreie Missionen zu gewährleisten. 
Die Verifikation der Missionspläne basiert nur auf unvollständigen und statischen Annahmen über die Umgebungsbedingungen, sodass trotz der funktional logischen Abfolge der Missionen deren Erfolg nicht garantiert ist. Die Verifikation der Missionsausführung berücksichtigt dagegen stärker die Umwelt und die aktuellen Bedingungen. Die damit verbundenen Regeln sind jedoch nicht darauf ausgelegt, den Erfolg der Mission zu gewährleisten.

Doch selbst die Kombination der beiden Konzepte erfüllt die Aspekte der Verlässlichkeit nur bedingt.
Diese erfordert weitere Überlegungen zu den anderen, nicht angesprochenen Aspekten der Verlässlichkeit, wie z. B. der Wartbarkeit, wie in \cite{Avizienis04} definiert. Die Autoren bewerten eine einzelne, allumfassende Methode zur Gewährleistung der Verlässlichkeit für unzweckmäßig und empfehlen einen modularen, kombinierbaren Ansatz zur Verifizierung verschiedener Aspekte.

In dem gezeigten Anwendungsfall ergänzen sich beide vorgestellte Konzepte, sodass die Mission erfolgreich durchgeführt werden kann. Die Autoren befürworten die Entwicklung weiterer Konzepte, insbesondere für Fehlertoleranz und Fehlervorhersage, wie sie in \cite{Avizienis04} definiert sind. Wenn solche Konzepte auf modularer Basis entwickelt werden, können sie leicht zu größeren Konstrukten kombiniert werden, was einen ganzheitlichen Ansatz für die Verlässlichkeit von autonomen Ux\kern-1pt Vs erleichtert.

\section*{Danksagung}
Diese Forschungsarbeit aus dem Projekt RIVA wird durch dtec.bw – Zentrum für Digitalisierungs- und Technologieforschung der Bundeswehr gefördert. dtec.bw wird von der Europäischen Union – NextGenerationEU finanziert.

\bibliographystyle{IEEEtran.bst}
\bibliography{bibliography}
\end{document}